\documentclass{article}

     \usepackage[final, nonatbib]{neurips_2020}

\usepackage[utf8]{inputenc} 
\usepackage[T1]{fontenc}    
\usepackage{url}            
\usepackage{booktabs}       
\usepackage{amsfonts}       
\usepackage{nicefrac}       
\usepackage{microtype}      

\usepackage{amsmath}
\usepackage{graphicx}
\usepackage{authblk}

\title{A step towards neural genome assembly}

\author[1,2]{\textbf{Lovro Vrček}}
\author[3]{\textbf{Petar Veličković}}
\author[1,2]{\textbf{Mile Šikić}}

\affil[1]{Genome Institute of Singapore, A*STAR, Singapore}
\affil[2]{Faculty of Electrical Engineering and Computing, University of Zagreb, Croatia}
\affil[3]{DeepMind \authorcr \tt \{vrcekl, miles\}@gis.a-star.edu.sg }

\begin{document}

\maketitle

\begin{abstract}
    \textit{De novo} genome assembly focuses on finding connections between a vast amount of short sequences in order to reconstruct the original genome. The central problem of genome assembly could be described as finding a Hamiltonian path through a large directed graph with a constraint that an unknown number of nodes and edges should be avoided. However, due to local structures in the graph and biological features, the problem can be reduced to graph simplification, which includes removal of redundant information. Motivated by recent advancements in graph representation learning and neural execution of algorithms, in this work we train the MPNN model with max-aggregator to execute several algorithms for graph simplification. We show that the algorithms were learned successfully and can be scaled to graphs of sizes up to 20 times larger than the ones used in training. We also test on graphs obtained from real-world genomic data---that of a lambda phage and E. coli.
\end{abstract}

\section{Introduction}

Recently, we have witnessed a wide adoption of machine learning in graph representations and approaches to problem-solving on graph-structured data. In the center of this research are usually graph neural networks (GNNs) \cite{scarselli2008graph, gilmer2017neural} which are applied to various problems on graphs, such as finding the shortest path \cite{graves2016hybrid, xu2019can}, traveling salesman \cite{vinyals2015pointer, joshi2019efficient}, and bipartite matching \cite{georgiev2020neural}. Many of the earlier approaches, however, give the model complete freedom to determine how to obtain the solution from raw inputs, by using only the final algorithmic solutions as the supervision signal.

This was addressed by Veličković et al. \cite{velivckovic2019neural}, where the authors have provided explicit step-by-step guidance for learning algorithm execution. Moreover, they recognized that many of the classical algorithms share related subroutines, which enabled them not only to teach a model to execute several algorithms simultaneously, but also to demonstrate positive knowledge transfer between learning different algorithms.

While all the aforementioned research, and plenty of other work, made a significant breakthrough in the domain of machine learning on graph-structured problems, not a lot of work has yet been done on real-world systems. In this work, motivated by the contribution of Veličković et al. \cite{velivckovic2019neural}, application to bipartite matching by Georgiev and Li{\'o} \cite{georgiev2020neural}, and utilizing algorithmic reasoning in reinforcement learning \cite{deac2020xlvin}, we take a step in that direction and apply the neural graph algorithm execution approach on the problem of \textit{de novo} genome assembly, which to this day remains as one of the most difficult problems in bioinformatics. \textit{De novo} genome assembly aims at reconstructing the original genome sequence from hundreds of thousands of relatively short sequences called reads, without any knowledge of what the original sequence looks like. Ability to reconstruct genomes with high accuracy would have numerous applications in biology and precision medicine. In the ideal case, the problem is based on finding a Hamiltonian path through a complex directed graph, which is known to be an NP-complete problem. Due to long repetitive regions in genomes and artifacts that arise during sample preparation or sequencing, some nodes and edges should not be included in the final path. However, as a result of numerous reads covering the same region in the genome and biological knowledge, graphs can be simplified in several steps by removing redundant nodes and edges from them.

Here, we construct a model which can learn three of these simplification algorithms simultaneously and perform them in parallel, thus paving the way towards a purely neural execution of genome assembly. The code is available at \url{https://github.com/lvrcek/NeuralLayout}.

\section{Problem setup}

\subsection{Genome assembly}

After the reads are obtained from a sequencer, one of the most common approaches for \textit{de novo} genome assembly nowadays is called Overlap-Layout-Consensus (OLC) paradigm.

\begin{figure}
  \centering
  \includegraphics[width=0.9\linewidth]{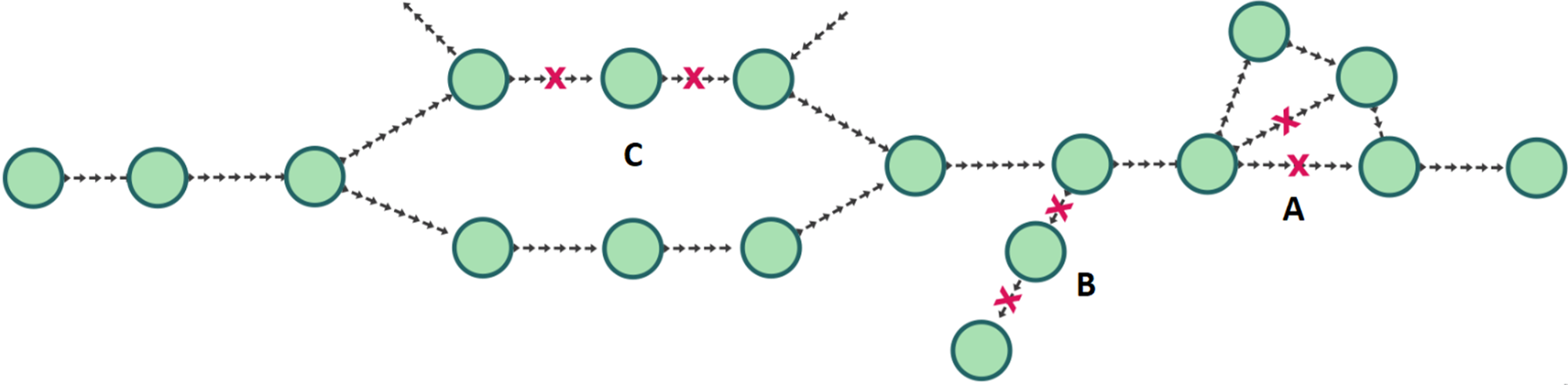}
  \caption{Example of structures in the assembly graph, before all the simplification steps. Letter \textbf{A} marks transitive edges, a short tip is marked with \textbf{B}, and a bubble which cannot be fully resolved is marked with \textbf{C}. Red crosses show which edges can be removed from the assembly graph.}
  \label{fig:graph}
\end{figure}

The overlap phase consists of mutually overlapping all the reads in the sample. Reads which are contained in the others are discarded immediately. From the rest of them an assembly graph is built, where nodes depict reads and edges depict the overlaps between them. Following this is the layout phase, where a path approximating the genome has to be found.
However, due to an abundance of reads some of them are redundant, and due to sequencing errors some overlaps are false. Thus, structures in the graph corresponding to such occurrences have to be detected and removed. OLC-based state-of-the-art assemblers such as Raven \cite{vaser2020raven} focus on three such structures---transitive edges, tips, and bubbles, which are the most common structures in the assembly graphs and can be seen in Figure \ref{fig:graph}. It is important to note that parts of the assembly graphs are paths even before the simplification, which makes the problem easier. Nevertheless, there exist regions of high complexity, which cannot be solved with the simplification algorithms described here.

After a few other processing steps, a path representing the assembly sequence is found. In the final phase, consensus, the assembly sequence is compared to the reads in the sample and polishing of the  sequence is performed.

The transitive reduction is the first simplification performed. Given a triplet of nodes $(n_i, n_j, n_k)$, edge $n_i \rightarrow n_k$ is transitive if there exist edges $n_i \rightarrow n_j$ and $n_j \rightarrow n_k$. Note that a transitive edge of one triplet can serve as a non-transitive in another triplet, and that the edge that "jumps over" more than one node is not transitive by this definition.

The second step is the removal of tips or dead ends---branches of several nodes that stem from the main chain and then cease abruptly, either due to sequencing errors or low sequencing coverage. They can be detected by running a depth-first search from the nodes with no outgoing edges and stopping once we find a node with multiple incoming or outgoing edges.

Finally, bubbles are the most complex structures that have to be removed. They are an arrangement of multiple paths connecting two nodes, where only one of these paths can be retained. However, nodes in those paths can be connected to other parts of the graph and removing them could result in a less accurate and more fragmented assembly sequence. This can be seen in Figure \ref{fig:graph} in the top path of the bubble, where not all the edges can be removed.

\subsection{Neural execution of simplification algorithms}

We model all three of the algorithms as graph traversal problems, where each of them has specific constraints on which edges can be traversed. The GNN model we use to learn these algorithms takes $T \in \mathbb{N}$ graphs $G = (V, E)$, where each node  $i \in V$ in the graph has associated node features $\vec{x}_i^{(t)} \in \mathbb{R}^{N_x}$ and each edge $(i, j)$ has associated edge features $\vec{e}_{ij}^{(t)} \in \mathbb{R}^{N_e}$. Here,  $N_x$ is the dimension of the node features, $N_e$ is the dimension of the edge features, and $t \in \{1, \dots T \}$ is a timestep in the execution as well as the index of the graph that is being processed. For each graph, a node-level predictions are made, simulating the next step in the algorithm execution.

The structure of the model follows the \textit{encode-process-decode} paradigm \cite{hamrick2018relational}, where encoder and decoder networks are specific for each algorithm, and the processor network is algorithm-agnostic. This enables us to learn multiple algorithms simultaneously. We also include the termination network, which predicts whether each algorithm should continue with the execution, as is done in \cite{velivckovic2019neural} where more information about the whole process can be found.

In short, this pipeline can be described with the following equations:
\begin{align}
    \vec{z}_i^{(t)} &= f_A \left( \vec{x}_i^{(t)}, \vec{h}_i^{(t-1)} \right),\\
    \mathbf{H}^{(t)} &= P \left( \mathbf{Z}^{(t)}, \mathbf{E}^{(t)} \right),\\
    \vec{y}_i^{(t)} &= g_A \left( \vec{z}_i^{(t)}, \vec{h}_i^{(t)} \right),\\
    \tau^{(t)} &= \sigma \left( T_A \left( \overline{\mathbf{H}^{(t)} } \right) \right).
\end{align}
Here, $f_A$ is the encoder network, $P$ is the processor network, $g_A$ is the decoder network, while the $T_A$ is the termination network. $\vec{z}_i^{(t)}$ and $\vec{h}_i^{(t)}$ are encoded and latent features for node $i$ at timestep $t$, while $\mathbf{H}^{(t)}$, $\mathbf{Z}^{(t)}$ and $\mathbf{E}^{(t)}$ represent matrices of all latent features, encoded features, and edge features, respectively. The predictions made by the network are represented by $\vec{y}_i^{(t)}$, and the predictions for termination by $\tau^{(t)}$. The input to the termination network is the mean latent state across all the nodes---effectively giving the latent state of the whole graph. Since we want to simulate algorithms as accurately as possible the next step of a deterministic algorithm is given at each step of neural execution as a supervision signal, thus training the model by teacher forcing.

\section{Results and discussion}

First, we have learned our model to execute all three of the mentioned algorithms separately. For training the model on each of them, 100 graphs had been generated. Each graph consisted of a path graph that is 50 nodes long, with additional structures we want the network to recognize added to that path graph. The positions of such structures and their sizes have been randomly generated for each graph. Testing was performed on graphs that were the same size, but also 2, 4, 8, and 20 times larger, which can be seen in Table \ref{table:res1}.

All the algorithm-dependent networks in our model are simple linear projections. For processor network we use a message-passing neural network (MPNN) with a max-aggregator. This architecture which was shown to have better performance than MPNNs with other types of aggregators and graph attention networks (GAT) \cite{velivckovic2017graph}. Message function used in the MPNN is also linear projections and update function is linear projection followed by a gated recurrent unit (GRU) \cite{cho2014learning}. We use Adam optimizer \cite{kingma2014adam} with a learning rate of 1e-5, and early stopping with the patience of 10 epochs. The latent dimension of $K = 32$ was used both for latent features and encoded node and edge features.

\begin{table}
  \caption{Scaling of algorithm execution for isolated learning of algorithms.}
  \label{table:res1}
  \centering
  \begin{tabular}{llllll}
    \toprule
    & \multicolumn{5}{c}{\textbf{Scaling}}                   \\
    \cmidrule(r){2-6}
    \textbf{Algorithm} & 1x & 2x & 4x & 8x & 20x\\
    \midrule
    Transitive removal     & 98.10\% & 99.00\%  & 99.52\% & 99.76\% & 99.91\%   \\
    Tips trimming          & 98.05\% & 98.96\%  & 99.49\% & 99.70\% & 99.87\%   \\
    Bubble popping         & 98.16\% & 99.03\%  & 99.53\% & 99.77\% & 99.90\%   \\
    \bottomrule
  \end{tabular}
\end{table}

After proving the network can learn all three of the simplification algorithms, we train the network to execute them simultaneously. The obtained accuracy is similar to the case of isolated learning for all three algorithms, from which we conclude that the network has successfully learned to execute them in parallel. Results can be seen in Table \ref{table:res2}.

\begin{table}
  \caption{Scaling of algorithm execution for parallel learning of algorithms.}
  \label{table:res2}
  \centering
  \begin{tabular}{llllll}
    \toprule
    & \multicolumn{5}{c}{\textbf{Scaling}}                   \\
    \cmidrule(r){2-6}
    \textbf{Algorithm} & 1x & 2x & 4x & 8x & 20x\\
    \midrule
    Transitive removal     & 98.21\% & 99.07\%  & 99.50\% & 99.89\% & 99.92\%   \\
    Tips trimming          & 98.45\% & 99.11\%  & 99.46\% & 99.76\% & 99.89\%   \\
    Bubble popping         & 98.17\% & 99.02\%  & 99.51\% & 99.78\% & 99.90\%   \\
    \bottomrule
  \end{tabular}
\end{table}

Finally, we test the parallel execution model on real-world data---we use lambda phage (\textit{Escherichia virus Lambda}) and \textit{Eschericia coli} genomic data sequenced with Oxford Nanopore Technologies sequencer. The assembly graph was produced with Raven assembler \cite{vaser2020raven}. We choose lambda phage because the graphs produced are small and convenient for more in-depth analysis, whereas E. coli is larger and gives better insight into how our model would generalize to eukaryotic genomes. Specifically, lambda phage graph consists of only 60 nodes, whereas E. coli graph consists of around 3000 nodes. The results can be seen in Table \ref{table:lambda}.

\begin{table}[h]
  \caption{Parallel algorithm execution on the assembly graph of lambda phage.}
  \label{table:lambda}
  \centering
  \begin{tabular}{cccc}
    \toprule
    & Transitive removal & Tips trimming & Bubble popping  \\
    \midrule
    Lambda phage & 98.04\% & 93.33\%  & 97.47\%    \\
    E. coli & 99.67\% & 98.84\%  & 99.26\%    \\
    \bottomrule
  \end{tabular}
\end{table}

The results on lambda phage are somewhat lower than on the generated graphs, especially for the trimming of dead ends. This can mostly be contributed to the fact that structures found in real-world assembly graphs are often significantly more complex than the structures found in the graphs generated for the training purposes. While the network was able to learn how to deal with transitive edges, dead ends, and bubbles, it was not exposed to structures which cannot be solved by the algorithms described here. Since the lambda phage graph is small, a single such structure can be detrimental to achieving high accuracy. On the other hand, E. coli graphs are large compared to the others used in this work. We have already shown that the model achieves higher accuracy on larger graphs which is due to parts of graphs being paths without junctions. The same is true for E. coli graph, which makes some predictions easy; thus, the accuracy is considerably higher than in the case of lambda phage.

Our next step is to generalize the model to solve complex structures which often cannot be solved exactly. Most of the state-of-the-art methods solve this part by manually set parameters of heuristic algorithms usually overfitting to several known model organisms, such as human. A neural model trained on complex and diverse dataset could simplify the genome assembly for less-known species. This urges us to train models on such dataset, which we leave to future work.

\section{Conclusion}
In this work we have successfully trained a neural model to execute three algorithms used for the graph simplification in \textit{de novo} genome assembly, which remains one of the most difficult algorithmic problems to this day. Although the learned algorithms are simple, they are an important part of the graph simplification phase in the assembly process. Furthermore, we prove this method is suitable for this type of algorithms and problems. After showing that the model is able to perform on graphs multiple times larger than the ones in the training set, we also show that it produces accurate predictions on real-world data. We believe this is a step towards automating the process of genome assembly, which will reduce the need for heuristics and hand-picked parameters.

\newpage

\begin{ack}
This work was performed at Faculty of Electrical Engineering and Computing, University of Zagreb, and Genome Institute of Singapore, A*STAR, as a part of the ARAP program. It was also partially funded by the European Union through the European Regional Development Fund under the grant KK.01.1.1.01.0009 (DATACROSS) and has been supported in part by the Croatian Science Foundation under the project Single genome and metagenome assembly (IP-2018-01-5886) and “Young Researchers” Career Development Program.

\end{ack}



\end{document}